\documentclass[letterpaper, 10 pt, conference]{ieeeconf} 
\usepackage[utf8]{inputenc}
\IEEEoverridecommandlockouts                       
\usepackage{mathtools}
\usepackage{newtxmath,newtxtext}
\usepackage{ragged2e}
\usepackage[font=small,labelfont=bf]{caption}

\usepackage{authblk}
\usepackage{pgfplots}
\usepackage{todonotes}
\usepackage{subfig}
\usepackage{graphicx}
\usepackage{layouts}
\pgfplotsset{compat=newest} 
\pgfplotsset{plot coordinates/math parser=false} 
\title{\Large \bf
Design and Evaluation of the SoftSCREEN Capsule for Colonoscopy} 

\author{\large Vanni Consumi$^{1}$ \emph{Member, IEEE}}
\author{\large Lukas Lindenroth$^{2}$ \emph{Member, IEEE}}
\author{\large Jeref Merlin$^{1}$ \emph{Member, IEEE}}
\author{\large \\Danail Stoyanov$^{1}$ \emph{Senior Member, IEEE}} 
\author{\large Agostino Stilli$^{1}$ \emph{Member, IEEE}} 
\vspace{10pt}

\affil{\small\textit{${}^1$Wellcome/EPSRC Centre for Interventional and Surgical Sciences (WEISS), University College London (UK)}\\\textit{${}^2$King's College London (UK)}
\\\textit{vanni.consumi.20@ucl.ac.uk, lukas.lindenroth@kcl.ac.uk, jeref.merlin.18@ucl.ac.uk}
\\\textit{danail.stoyanov@ucl.ac.uk,
a.stilli@ucl.ac.uk}}

\begin{document}

\maketitle
\thispagestyle{empty}
\pagestyle{empty}

\begin{abstract}
Colonoscopy is considered the golden standard for cancer screening of the lower gastrointestinal (GI) tract, with screening programs all over the world considering lowering the recommended screening age. Nonetheless, conventional colonoscopy can cause discomfort to patients due to the forces occurring between colonoscopes and the walls of the colon. Robotic solutions have been proposed to reduce discomfort, and improve accessibility and image quality. Aiming at addressing the limitations of traditional and robotic colonoscopy, in this paper, we present the SoftSCREEN System – a novel Soft Shape-shifting Capsule Robot for Endoscopy based on Eversion Navigation. A plurality of tracks surrounds the body of the system. These tracks are driven by a single motor paired with a worm gear and evert from the internal rigid chassis, enabling full-body track-based navigation. Two inflatable toroidal chambers enclosing this rigid chassis and passing through the tracks, cause them to displace when inflated. This displacement can be used to regulate the contact with the surrounding wall, thus enabling traction control and adjustment of the overall system diameter to match the local lumen size. The design of the first tethered prototype at 2:1 scale of the SoftSCREEN system is presented in this work. The experimental results show efficient navigation capabilities for different lumen diameters and curvatures, paving the way for a novel robot capable of robust navigation and reliable control of the imaging, with potential for applications beyond colonoscopy, including gastroscopy and capsule endoscopy.
\end{abstract}

\section*{INTRODUCTION}
\justifying 
Colorectal cancer (CRC) is the third deadliest cancer-related disease worldwide \cite{WHO/EuropeCancer}.
Nowadays, colonoscopy is the golden standard procedure for colorectal screening although the use of the standard semi-flexible endoscopes \cite{Kohli2019HowWork} can cause discomfort for the patient due to the stretching of the colon wall when the scope advances \cite{Appleyard2000TheColonoscopy}.
In the past two decades, ingestible robotic capsules have expanded the possibilities for diagnosis and interventions achievable in the GI tract \cite{Ciuti2016FrontiersReview}.
Several active robotic colonoscopes, mostly tethered, have been proposed with the aim of inspecting the lower GI and overcoming the drawbacks of standard colonoscopy \cite{Ciuti2020FrontiersTechnologies}. One new concept prevalent in the field is the creation of self-propelled systems embedding cameras and tools able to navigate inside the intestine, minimising patient discomfort by reducing the interaction forces with the intestine lining \cite{Kassim2006LocomotionColonoscopy}. 
\begin{figure}
    \centering
    \includegraphics[width= \linewidth]{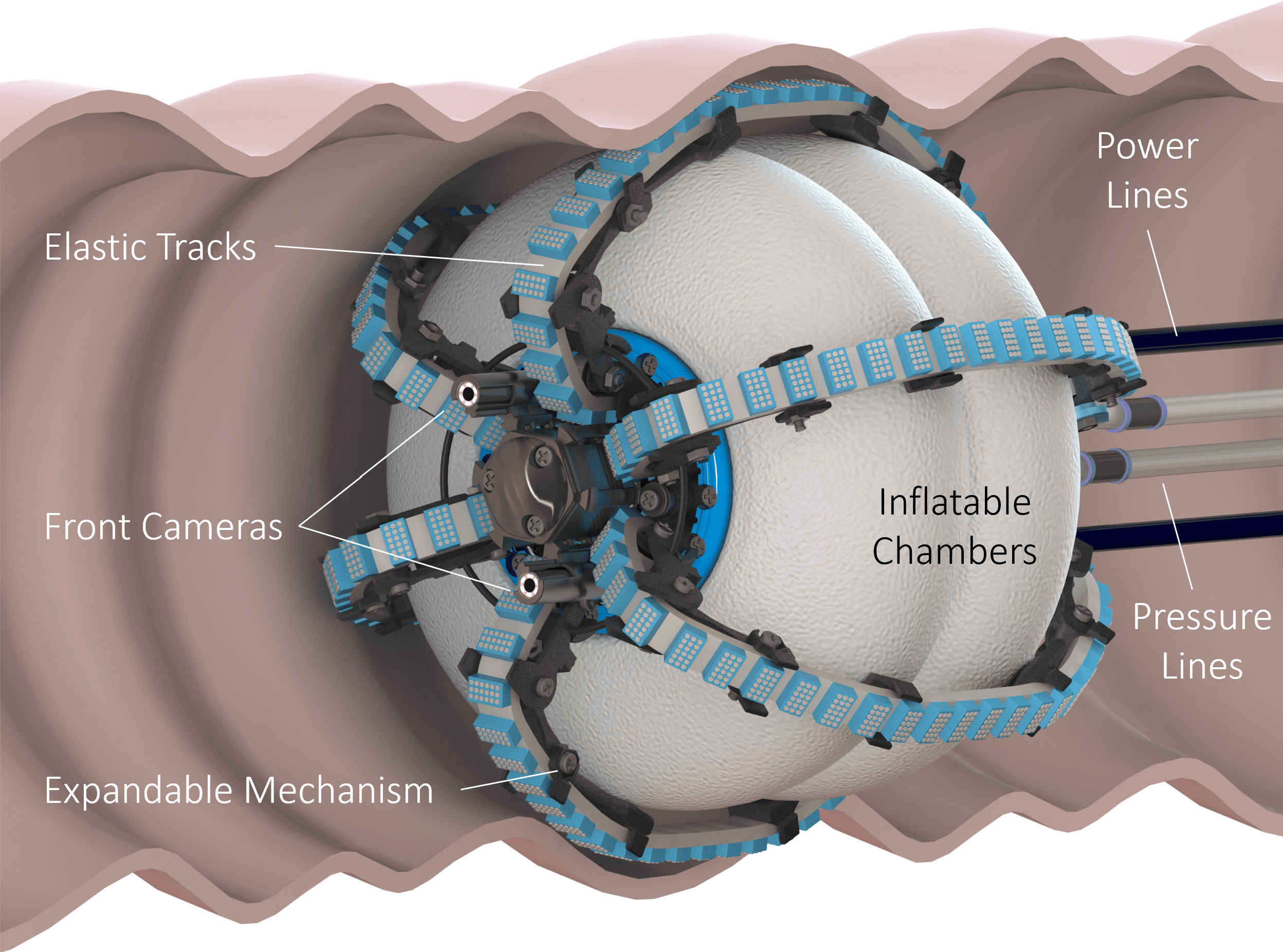}
    \caption{Render of the SoftSCREEN system navigating inside the colon.}
\end{figure}
Magnetically controlled capsules have also been proposed and are currently being extensively investigated. These systems navigate by being propelled by an external magnet and by controlling the orientation of the external magnetic field steering of the capsule within the lumen is achieved. Magnetic navigation of ingestible capsules showed comparable screening outcomes when compared with conventional gastroscopy in the clinical study presented in \cite{Liao2016AccuracyDiseases}. Tethered magnetic-driven systems for colonoscopy showed promising results in the first tract of the colon (sigmoid and descendent) but limited navigation capabilities beyond the splenic flexure as shown in porcine \textit{in vivo} test presented in  \cite{Valdastri2012MagneticColonoscopy}. The EU Project Endoo demonstrated pre-clinical feasibility for the exploration of the lower GI with magnetic capsules \cite{Verra2020Robotic-assistedCapsule}.
Researchers have also proposed bio-inspired locomotion strategies such as legged robots \cite{Stefanini2006ModelingEnvironment}, inch-worm robots \cite{Manfredi2019AColonoscopy} \cite{Wang2013AColonoscopy}, or locomotion by mean of wheels \cite{Norton2016RollerBall:Locomotion}, paddles \cite{Osawa2020Self-PropelledPaddles} or tracks \cite{Sliker2012SurgicalTreads} \cite{Formosa2020NovelColonoscope}. Typically, the use of tracks ensure larger contact areas than wheels, hence, more reliable traction; the grooves of the tracks can establish geometrical coupling with the folds of colon tissue that increases traction \cite{Wang2009ResearchCharacteristics}.\\
Track-based robotic colonoscopes such as \cite{Sliker2012SurgicalTreads} or \cite{Formosa2020NovelColonoscope} arrange tracks around a rigid chassis in a fixed configuration and, as such, cannot ensure full-body traction due to the widely changing diameter of the colon from the anus to the cecum. The flexible track-based systems presented in \cite{Kim2013ACaterpillars} and \cite{Kim2014TheShaft} proposed a design with some level of passive shape-adaptability to the navigated lumen enabled by the elasticity of the tracks, however, the effect of external forces, including gravity, can displace the tracks in undesired positions and negatively affect the locomotion. The system \cite{Kim2014TheShaft} exhibited sufficient locomotion in \textit{ex vivo} intestine tests but did not achieve full cecal intubation during \textit{in vivo} pig tests.
In the context of tethered capsules for colonoscopy, Ortega et al. \cite{OrtegaAlcaide2021Tether-colonDevices} highlighted how the friction of the tether on the colon walls increases with the number of flexures navigated. The overall resistance, evaluated up to 4 N, can result in the failure of cecal intubation. For this reason, the ability of an active colonoscopic system to control the interaction forces between its locomotion elements, being this legs, tracks or wheels, and the walls of the navigated lumen, is a crucial feature to achieve large propulsion forces.
Air inflating chambers made of silicone, such as those integrated in inch-worm robots \cite{Manfredi2019AColonoscopy}, can expand the diameter of the system with a level of compliance that can limit circumferential stress generated in the colon and associated with pain \cite{Loeve2013MechanicalAvoided}. 
To produce anisotropic deformations and guide the deformation in specific direction, inflatable chambers are designed with walls of heterogeneous thickness or with mechanical constraints \cite{Gorissen2017ElasticApplications}.\\
In this paper, we present the SoftSCREEN System – a Soft Shape-shifting Capsule Robot for Endoscopy based on Eversion Navigation. It employs inflatable chambers to create a wall-pressing mechanism for a series of elastic tracks arranged on the external surface of the robot. This combination of deformable elements ensures full-body navigation with a soft interaction with the walls of the navigated lumen, providing control on the traction force as well as on the overall shape of the system. The capability of the system to vary its shape (shape-shifting) also allows for the alignment of the front camera with the center of the lumen, ensuring optimal view during navigation. In this work, we propose the design of the system and we validate its two main functions, namely the locomotion and the shape-shifting capability, envisioning its applicability for colonoscopy. The detailed study here presented focuses on a 2:1 scale prototype of the robot.

\section*{MATERIALS AND METHODS}
\justifying 
A render of the design of the inflated SoftSCREEN system inside a colon is displayed in Fig. 1.
The robot is composed of a set of elastic toothed tracks, regularly spaced along the circumference to guarantee 360° contact with the walls of the navigated lumen, two inflatable toroidal chambers and a rigid cylindrical hollow chassis. Inside the chassis, a worm gear directly engages with the teeth of the tracks and it is put in rotation by an electric motor.
The force applied by the rotation of the worm gear causes the motion of the tracks inside the chassis, where longitudinal guides constrain them to move linearly so that they realise a continuous motion loop from the front to the back of the robot body or vice versa, according to the direction of rotation of the worm gear. As depicted in Fig. 1, the inflation of the chambers displaces and deforms elastically the tracks, putting them in contact with the walls of the lumen: the inflation level determines the normal force applied by the tracks on the walls as well as how many tracks are simultaneously in contact.
The internal smooth surface of each track slides on the external surfaces of two chambers, while the external toothed surface is in contact with the walls of the navigated lumen. 
The friction between the walls of the lumen and the toothed surface of the tracks generates the force that propels the system, and the direction of motion of the system is the same as the direction of motion of the tracks inside the chassis: the capsule is propelled forward when the tracks evert from its front and backward when the tracks evert from its back.
The system has a single degree of freedom (DOF) for the navigation, however, we will see how the inflation of the two chambers can provide additional DOFs to the system. 
\begin{figure}
\centering
\includegraphics[width= \linewidth]{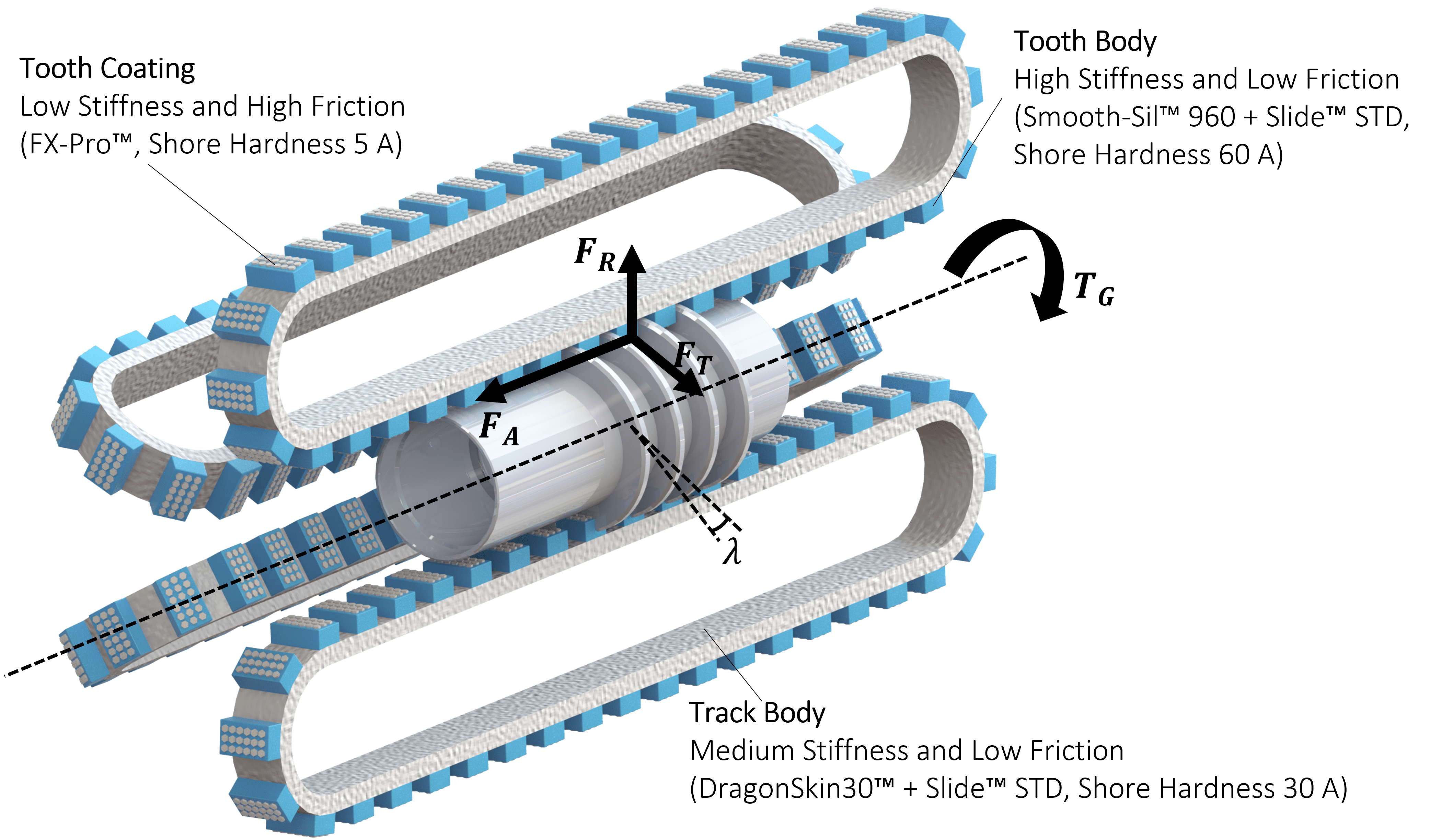}
\caption{Overview of the forces transmitted from the worm gear to the elastic tracks and material selection for the tracks: a 30A silicone for the elastic body of the track, a 60A silicone for the teeth of the track for a sturdy mechanical transmission with the worm gear, a 5A silicone coating layer with a micro pattern on the top of each tooth to locally increase the friction coefficient.}
\end{figure}
Due to its functioning, relative sliding motion between the tracks and the other parts of the system (worm gear, chassis, silicone chambers) occurs. Therefore, the efficiency of this transmission can be increased by reducing the friction between sliding surfaces, e.g. adopting low-frictional materials and/or reducing the contact pressure. 
In this section, the design of the robot is introduced in details. Starting from a 2:1 scale system, we derived the design and developed the prototype, considering the scalability of all the components, manufacturing processes, and the dimensions of the components used in the fabrication.  
\subsection{Worm gear mechanism}
The propulsion force that enables navigation is obtained through a set of toothed elastic tracks made of multiple silicones driven by an internal worm gear mechanism. In Fig. 2 this transmission is presented in details, showing the different types of silicones used for the different elements of the tracks: high-stiffness body of each tooth to ensure reliable force transmission between the worm gear and teeth, a low-stiffness silicone coating on the top part of the teeth to ensure high frictional slippage when this surface engage with the walls of the navigated lumen; a medium-stiffness silicone for the elastic body of the track to deform upon inflation of the chambers. In the ideal scenario, where the external part of the tracks in steady contact with the navigated wall with no slippage, the relation between the robot linear speed $v_{robot}$ in the direction of the main axis of symmetry of the system, the rotational speed of the worm gear $\omega$ expressed in $rad/s$  and the pitch $p$ can be expressed as:
$$
v_{robot}=\frac{\omega p}{2\pi} \eqno{(1)}
$$
The use of multiple silicone materials for all the locomotion members provides a frictional interface with the colon walls while also ensuring the overall compliance of the system, guaranteeing the safety of the lining upon navigation as shown in the scenario in Fig 1. 
Referring to Fig. 2, the tangential force $F_T$ of the worm gear depends on the torque of the internal gearhead (motor+gearbox) $T_G$ and the pitch diameter of the worm gear $D_W$:
$$
F_T = \frac{2T_G}{D_W} \eqno{(2)}
$$
The worm gear exerts an axial force $F_A$ on the elastic tracks depending on the lead angle $\lambda$ of the worm gear mechanism:
$$
F_A = \frac{F_T}{\tan\lambda} \eqno{(3)}
$$
In addition, a radial force $F_R$ function of the pressure angle of the tooth of the gear is also present and it tends to separate the worm gear from the tracks when the two enter in contact, and it is defined as below:
$$
F_R = F_A\tan\phi \eqno{(4)}
$$
The torque of the gearhead is function of the motor torque $T_M$, the gearbox ratio $i$ and the efficiency $\eta$ of the gearbox as follows:
$$
T_G = T_M  i  \eta \eqno{(5)}
$$
The pitch of the worm gear $p$ is selected as 6 mm, the pitch diameter $D_W$ is 18mm, the lead angle $\lambda$ of 5.5° has been chosen to obtain large axial component of force in Eq. 3 and a small tangential component on the soft tracks, as it would result in larger tangential forces on the tracks against the lateral walls of the internal groves guiding the tracks on the internal surface of the chassis. The pressure angle of the gear of the worm gear $\phi$ is 0° (squared gear) to prevent the creation of the radial force components of Eq. (4).\\
Due to the extensive friction contact that the use of a worm gear transmission entails, the frictional loss between the elements of the transmission can be high. On the material side, we combined soft and hard silicones to ensure teeth stiffness and reduced friction between surfaces in sliding contact by using tension diffuser additives (SLIDE™ STD, Smooth-On Inc.). To reduce large contact areas in the transmission, each track engages the worm gear with 4 teeth only, as represented in Fig. 2 along with clear details about the different materials and additives used for the different deformable elements.
\subsection{Finite Element Analysis}
\begin{figure}
\includegraphics[width= \linewidth]{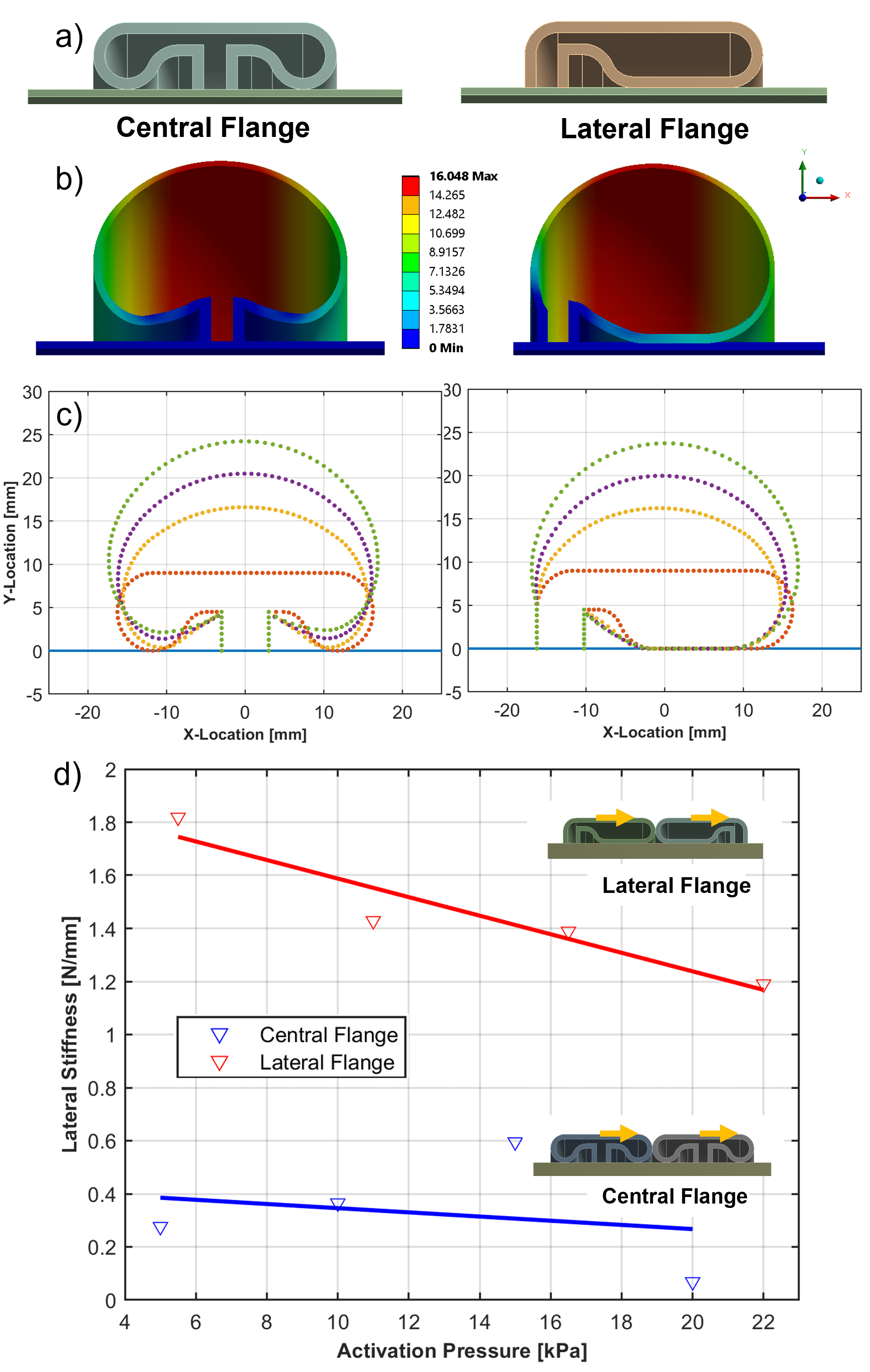}
\caption{a) Left: central flange (CF) profile. Right: lateral flange profile (LF). b) Total deformation from simulation (16 mm maximum). c) Location of the nodes at 4 different time-steps. d) Trend of the lateral stiffness for the configuration with 2 chambers, obtained from simulation of 4 levels of inflation and by applying an external force of the upper surface.}
\end{figure}
For the design of the inflatable chamber, we considered the following aspects: in an ideal scenario, the chambers should exhibit only radial deformation upon inflation which is useful to adjust the overall diameter of the system. Axial deformation should be minimised as it only causes additional mechanical stress on the tracks. Furthermore, chambers should be made of silicone by injection moulding to ensure reliable and repeatable manufacturing and symmetric inflation. The design must also allow for a robust and air-tight coupling with the chassis of the robot. The material should provide a low frictional interface with the tracks sliding on the external surface of the chambers. The chambers should retain their shape in both inflated and not inflated states, while tracks are sliding on their lateral surfaces, hence, they should exhibit high axial stiffness to contrast the shear force at the interface with the tracks. \\
Considering this, an optimisation study based on Finite Element Analysis (FEA) has been conducted. The study is based on ANSYS™ 2019 (ANSYS Inc., Canonsburg, PA, US) and it is described in this section. For our application, the design of the silicone chambers includes two circular flanges to airtightly secure them on the lateral cylindrical surface of the chassis. Thus, sufficient space beside the flanges was considered to place a  rigid ring with passing screws for fixture. Given this, we have considered the two toroidal designs shown in Fig. 3-a. These configurations are named Central Flange (CF) and Lateral Flange (LF) and they occupy an identical volume around the chassis but they differ in the location of the flange. 
We then simulated the free inflation of these toroidal bodies to evaluate their deformation. The chambers are constrained by the flanges, whilst pressure is applied inside them; a single circular section of the chamber (60\textdegree) is modelled in the software, but the entire toroidal behaviour is captured by imposing cyclic symmetry condition. While the chassis in the simulation is made of steel, as they will be in the final system, the chambers are made of DragonSkin20™ (Smooth-On Inc., Macungie, PA, US), modelled as 1st- order Ogden hyper-elastic material which uniaxial tensile test values were obtained from \cite{Marechal2020TowardRobotics}. Frictionless contacts were assumed in the simulations. \\
Due to the limitation of the FEA software used for this simulation study, the maximum radial deformation achievable with both models, hence, useful to compare their performance, is around 16 mm radial (Fig 3-b). The two profiles exhibit comparable internal stress, with a maximum stress of 0.61 MPa for the LF configuration and 0.49 MPa for the CF one. The activation pressure, which is the pressure level in the chambers, is also comparable, as well as low axial deformation (Fig. 3-c). However, simulation results showed significant differences in the behavior upon shear load: if we define the axial stiffness of the chambers, $k_a=F/d$, as the shear force applied on the external surface of the chambers divided by the maximum lateral displacement measured in the simulation, the configuration using two LF chambers showed an axial stiffness greater than 4 times the axial stiffness measured for the same inflation level using CF chambers (Fig 3-d). The axial stiffness of the chambers showed to decrease with the increment of the internal pressure under the same shear force, which was selected as 0.5 N for these simulations. The axial stiffness exhibited by the LF profile strongly depends on the contact occurring almost immediately upon inflation between the chassis and the chamber (Fig 3-c) that enables the LF chamber to anchor around the chassis, with a maximum contact pressure of 40 kPa according to the data collected in the simulation, and maintain the shape under shear force on the external surface. As such, LF configuration was selected for the subsequent FEA study.\\
\begin{figure}
\includegraphics[width= \linewidth]{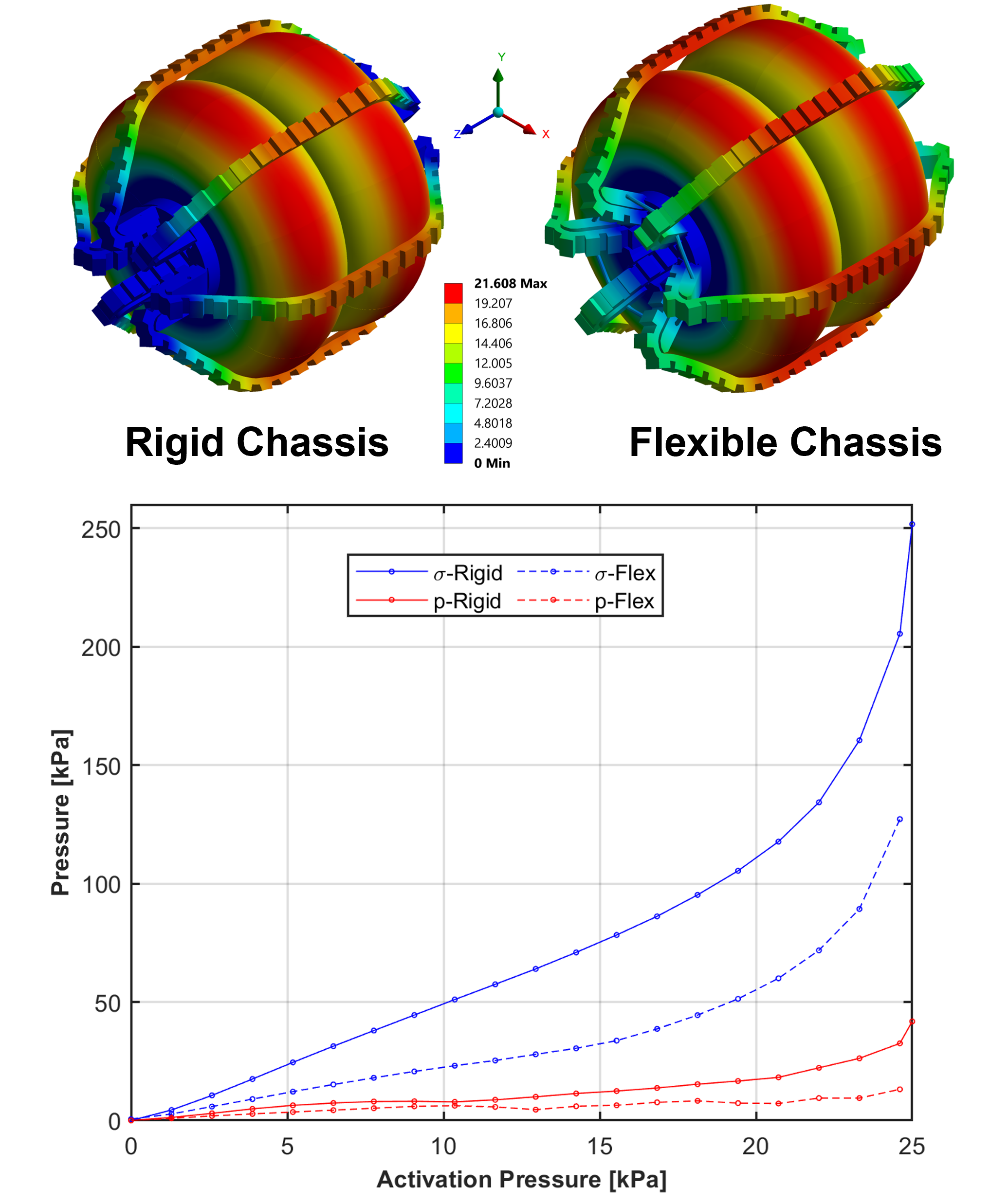}
\caption{Internal stress in the track and contact pressure between track and chamber for the system with Rigid Chassis and Flexible Chassis (maximum total deformation 21 mm).}
\end{figure}
Regardless of the flange position, the volumetric expansion of the chambers upon inflation will always cause strain on the tracks displaced by their deformation, with an overall increase of the stress of the tracks and of the tracks-chambers contact pressure. For this reason, a circular section (with cyclic symmetry) of the system composed of the two chambers (made of DragonSkin20™), tracks (made of DragonSkin30™ as for the tracks’ body in Fig. 2) and the chassis (made of steel) is simulated both considering a fixed and a reconfigurable chassis. In the reconfigurable version, to reduce the stress on the tracks upon inflation of the chambers, the two distal ends of the internal guides can deploy radially by rotating around a passive joint at the end of the fixed portion of the chassis to which the chambers are connected as depicted in the top of Fig 4, where the two geometries are presented. In this design, each deployable track guide can return to its rest configuration thanks to an elastic cable encircling all these elements on both ends of the capsule. This elastic cable has been considered in the simulation to be made of DragonSkin20™. As shown in Fig. 4, simulation results at a maximum radial deformation of the system of 21 mm, demonstrate a significant reduction of the internal stress in the tracks as well as of the contact pressure between tracks and chambers by almost 50\%, suggesting a reduced risk of failure of the tracks and less sliding friction occurring during the everting motion of the tracks.\\
As a result of this simulation study, the reconfigurable design combined with the lateral-flanges configuration has been selected for the development of the proof-of-concept system presented in the next section.
\subsection{Fabrication of the prototype}
\begin{figure*}
\includegraphics[width= \linewidth]{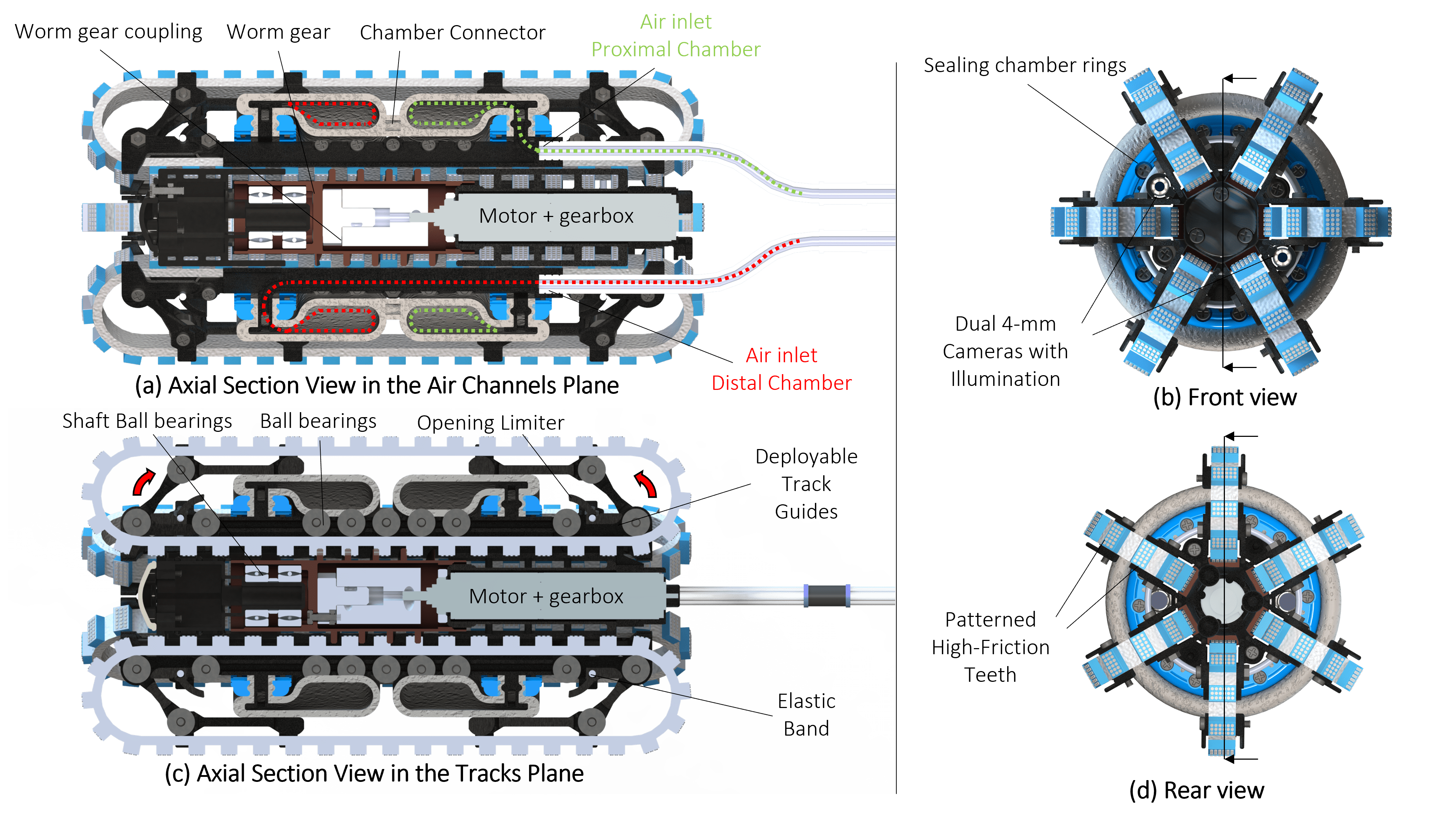}
\caption{Front view, section of the air channels, rear view, section of track lanes. Red arrows in axial section highlight the rotation of the deployable links. The two chambers are secured to the chassis using two sets of two gaskets fixed to the flanges with bolts.
}
\end{figure*}
\begin{figure*}
\centering
\includegraphics[width= \linewidth]{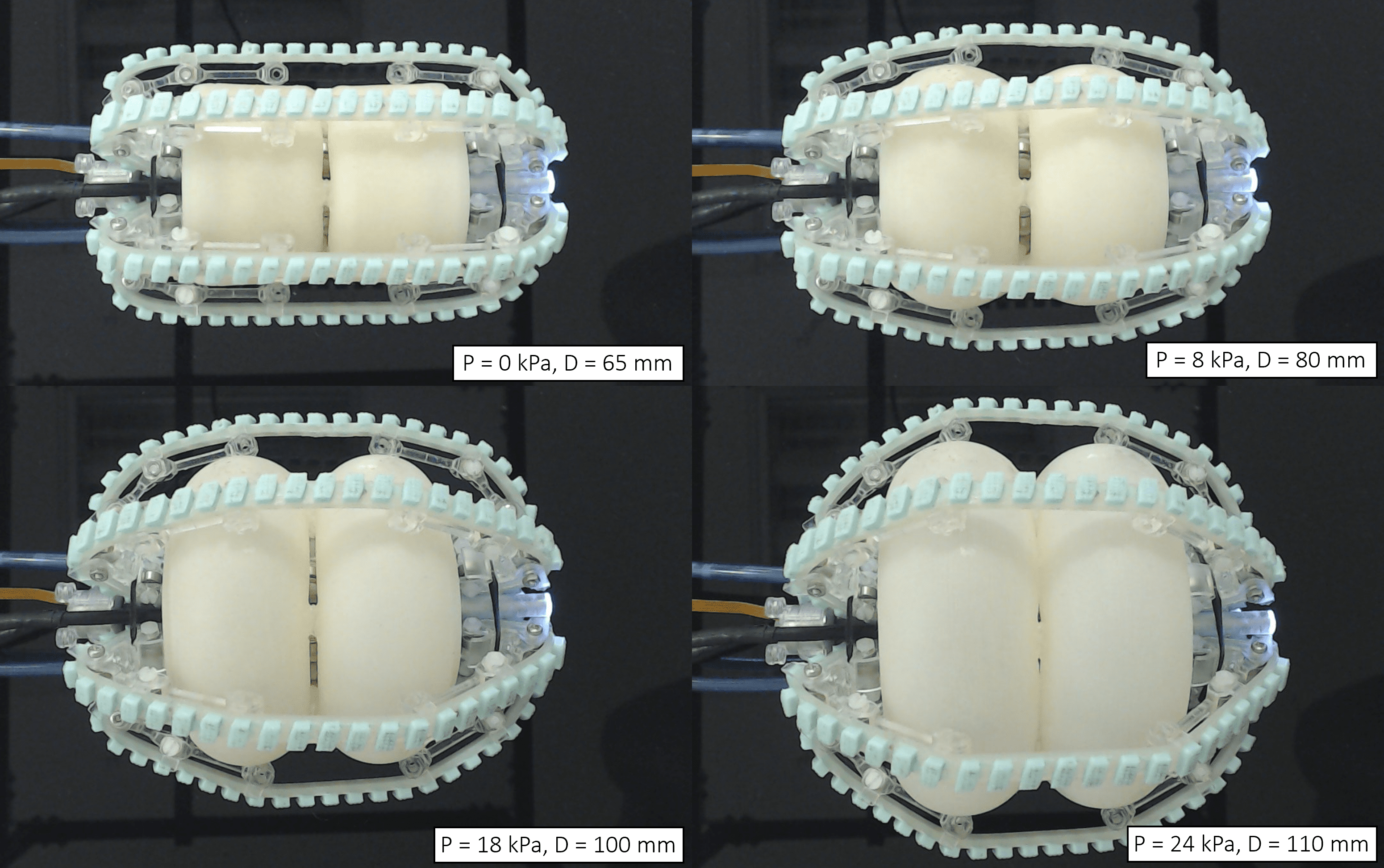}
\caption{Four stages of inflation of the silicone chambers with pressure supplied from external regulators. The expandable mechanism, composed of three deployable segments for both sides of the chassis and for each track, opens in accordance with the radial deformation and protects the elastic tracks from lateral displacements.}
\end{figure*}
\begin{figure}
\centering
\includegraphics[width= \linewidth]{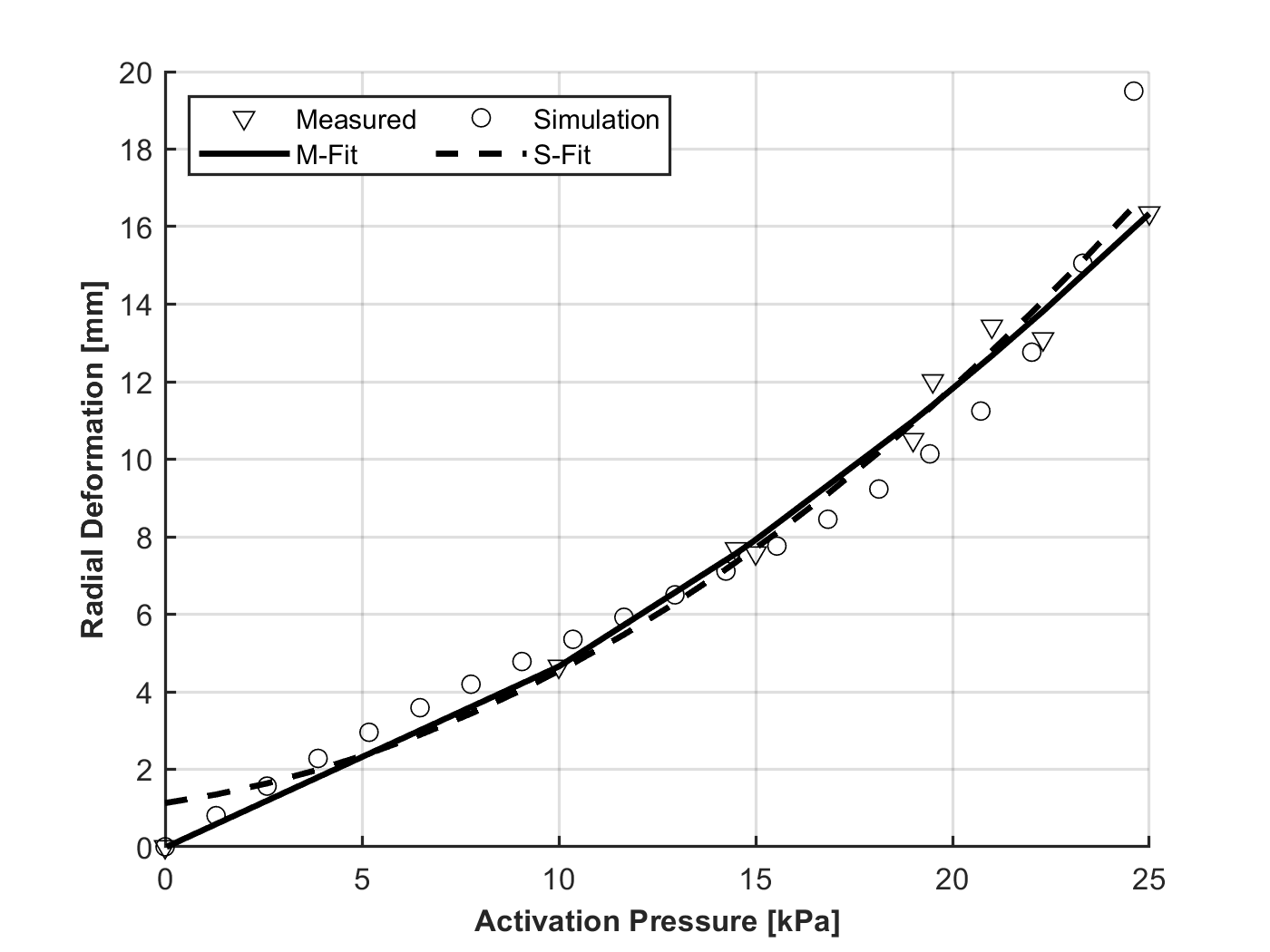}
\caption{Maximum radial deformation of the chambers measured both experimentally and from simulation, with fitting curves.}
\end{figure}
An overview of the design of the proof-of-concept system is presented in Fig. 1 in perspective view navigating the colon and in more details in front, rear and two longitudinal section views in Fig. 5. The robot’s core is composed of a central, rigid, cylindrical body, which encases the geared motor (a DC brushless motor 315170 with 256:1 ratio gearbox and rotatory encoder from Maxon Motor™, Sashseln, Switzerland) and the worm gear paired with it (Fig. 5-a). The two toroidal inflatable chambers are made of silicone membranes with an LF profile, secured with screws on two flanges on the chassis of the robot. The chambers can be independently inflated via two air channels routed through the chassis and then connected with a soft tethered line to a pressure supply. As shown in Fig 5-b, ball bearings are placed along each internal track guide with the aim of reducing the sliding friction occuring between the non-toothed surface of the tracks and the surface of the guides inside the chassis. Furthermore, since a compliant chassis is expected to reduce stress and contact pressure of the tracks as shown in the preliminary FEA, the terminal parts of the internal guides embedded in the chassis have been designed to provide a certain degree of compliance to the deformation of the chambers and of the tracks by means of twelve deployable track guides mounted at the six entry and exit points of the tracks as shown in Fig. 5-c. This deployable chain mechanism is constructed of 3D-printed elements and three hinges, made of a ball bearing as pivot, that settle between the chambers and the tracks, with the aim of mechanically limiting the lateral displacement of the tracks. This deployable passive system is activated by the radial deformation of the chambers occurring upon their inflation. When the chambers are in a not inflated state, this mechanism stays closed following the straight profile of the internal portion of the guide thanks to two elastic cables each encircling six of the twelve deployable elements. An opening limiter that stops the maximum opening of these elements to an angle of 60\textdegree from the resting position is used to prevent excessive contact with the inflating chambers and to avoid undesired deformations. As shown in Fig. 1 and Fig. 5, the proof-of-concept system embeds two endoscopic cameras (3.5 mm diameter) with LED lighting to provide a full frontal field of view. The inflatable mechanism ensures that is always possible to match the local lumen navigated while also ensuring that the cameras, as well as the central axis of the system, are aligned with the central axis of the navigated lumen. Rigid components are 3D printed with Clear Resin (Formlabs™, Somerville, MA, US) and silicone parts are fabricated using injection moulding.
Tracks are manufactured by over-moulding to merge three different silicones as shown in Fig. 2. Silicone chambers are made of DragonSkin20™ (Smooth-On Inc., Macungie, PA, US) with the SLIDE™ STD  additive from the same brand to reduce the surface friction. The velocity and level of inflation of the robot is achieved by controlling the speed of the motor via the motor driver DEC Module 24/2 connected to DEC Module Evaluation Board (Maxon Motor™, Sashseln, Switzerland), and through two VPPX pressure regulators (FESTO™, Esslingen am Neckar, Germany) controlling the inflation level of the two chambers. At the current stage, the robot is controlled in open loop for both position and velocity. The values of the pressure of the air measured by the regulators in each chamber are captured and displayed via a microcontroller (Arduino UNO board).

\section*{RESULTS}
\justifying 
This section presents the testing results of the fully assembled SoftSCREEN prototype depicted in Fig. 6.
\subsection{Inflation of the system}
In the proposed design, the robot embeds two front cameras for visual inspection of the inside of the lumen. Although the location of the cameras is fixed in the chassis, if consistent inflation is delivered to both chambers, the diameter of the robot will match the diameter of the lumen in which it is navigating, and it will self-centre the field of view of the cameras based on this, thus optimising cameras orientation for image acquisition during a colonoscopy procedure. As shown in Fig. 7, the maximum radial displacement exhibited by the chambers was measured by processing images of the prototype (placed on a stand and free to inflate) in MATLAB™ 2022 (MathWorks, Portola Valley, CA, US). We then compared it with the maximum radial deformation of the chambers evaluated in the FEA environment in the case of the Flexible Chassis system; the fitting curves that interpolate the measured points show comparable behaviour between the measurement and the simulation. 
Moreover, since the pressure lines of the two chambers are independent by design, the user can pressurise the two chambers at different pressure levels and combine the effect of differential pressurization of the two chambers and the effect of gravity acting on the capsule to enable an additional DOF for the robot, that will be able to tilt with respect to the longitudinal axis of the lumen navigated. The range of tilting angle achieved by the capsule with respect to the initial, non-inflated, position was determined with the robot inside a transparent rigid pipe with 94 mm diameter, and it was quantified in a range of ±10°.
\begin{figure*}
\centering
\includegraphics[width= \linewidth]{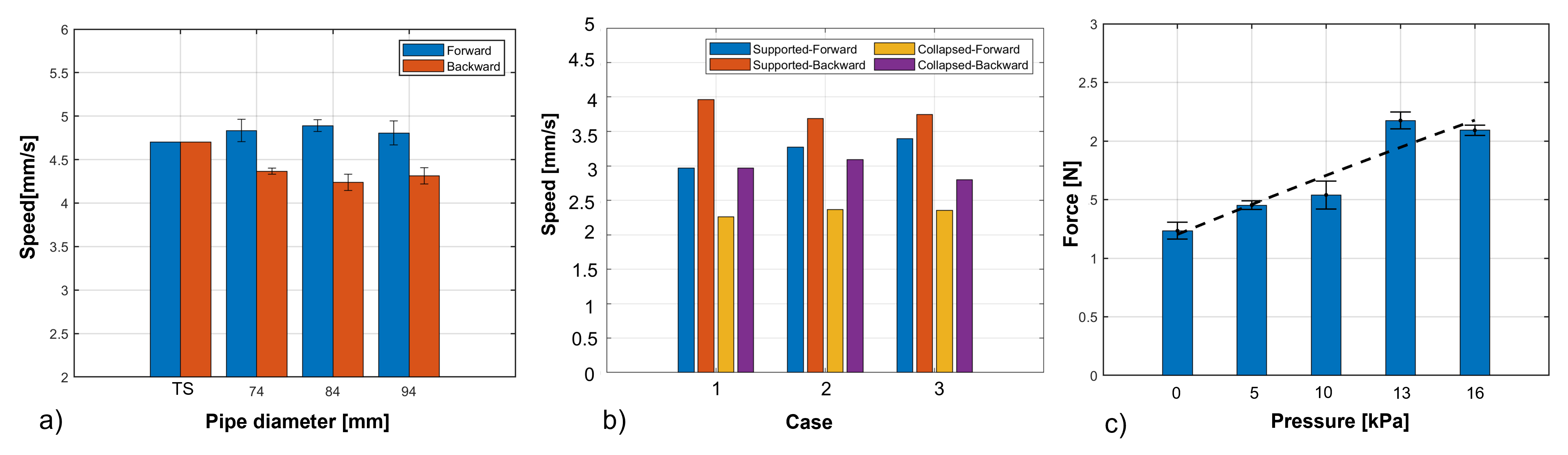}
\caption{a) Speed of the robot inside rigid pipe at different diameters with standard deviation and theoretical speed (TS). b) Speed of the robot in the Supported and Collapsed phantom, and at three different cases of chambers’ inflation; average speed of three consecutive tests is reported.
c) Traction force exerted by the robot: the force is evaluated at five different inflation pressures of the chambers (0 kPa, 5 kPa, 10 kPa, 13kPa and 16 kPa); average of the peaks, standard deviation and trend are reported.}
\end{figure*}
\subsection{Locomotion experiments}
The locomotion of the robot is preliminarily evaluated inside a set of 3 acrylic pipes of different internal diameters (74 mm, 84 mm, 94 mm) with the robot measuring around 65 mm in the non-inflated configuration. For this test, the robot is inflated to match the internal diameter of the pipe so that at least 5 out of 6 tracks are visibly in contact with the internal wall of the pipe before the beginning of the test. The speed of the robot is measured in forward and backward directions inside the pipe placed horizontally, over five consecutive tests for each pipe diameter, when the motor runs at its maximum speed (circa 12000 RPM). The speeds of the robot are presented in Fig. 8-A with respect to the theoretical speed value obtained from Eq. (1). This test demonstrated the robot is able of navigating inside rigid pipes, re-configuring its shape accordingly with the pipe to keep the tracks in contact with the wall without major effects on the speed.\\
Fig. 8-B shows the estimated speed of the robot inside a tubular, soft phantom. The phantom used in this work is composed of three parts (two straight segments and one 90° elbow, total length of circa 600 mm) made of silicone parts (DragonSkin FX-Pro™ - Smooth-On Inc., Macungie, PA, US) cast with an open mold using wavy tubular profiles (circa 85 mm diameter to represent almost a 2:1 scale colon). The mould is then reversed in order to obtain an internal rough surface; the mould of the turning section is a 90° elbow pipe with a radius of curvature of around 75 mm on the midline.
Two different scenarios for the phantom are considered.
In the first case (called Supported Phantom), four circular rigid supports are placed at the entrance and the exit of the phantom, and before and after the 90° elbow, to sustain the phantom to replicate the clinical scenario of an insufflated colon. 
In the second case (called Collapsed Phantom), the phantom is not held by any supports and is fully collapsed on the robot. 
For each of these two cases (Supported and Collapsed), three different inflation rates of the chambers of the capsule were tested inside the phantom: with reference to Fig 8-B, we distinguished a case where no pressure is applied to any chambers (Case 1 - No Inflation), a case in which a pressure of 10 kPa is applied to both chambers (Case 2 - Medium Inflation), and a case in which a 16 kPa is applied to both chambers (Case 3 - High Inflation), where all the six tracks were observed to be fully in contact with the phantom wall in the case of Collapsed Phantom, and the membrane of the phantom was subjected to notable radial stretch.
Moreover, silicone lubricant (WD-40, WD-40 Company, San Diego, CA, US) is applied inside the membrane at the beginning of the experiment to reduce the frictional coefficient of the silicone.
For each case study, the speed of the robot is measured from the entrance to the final exit of the membrane by video recording. 
The robot showed to be able to navigate the phantom in both Sustained and Collapsed scenarios (Fig. 8-b) and faster navigation is performed in the Supported case, with a peak of 3.9±0.17 mm/s backward speed for the non-inflated case, and a maximum of 3.4±0.15 mm/s forward speed. Generally slower motion is observed in the collapsed phantom, with a stable forward speed of around 2.35±0.12 mm/s forward speed and a backward maximum of 3.1±0.25 mm/s.
During the tests, the user gently conducts the power/air tethers of the robot by hand.\\
In Fig. 8-C, the maximum traction force exerted by the robot inside the silicone membrane is displayed. Traction force is crucial to achieve robust navigation inside the colon and necessary to overcome the resistance force arising by the tether during the cecal intubation. For this experiment, the robot is positioned inside the silicone phantom in the Collapsed Phantom scenario and the maximum propulsion force is measured as function of the chambers' pressure, by a load cell connected to the back of the robot with an extendable wire and a metal spring. For higher inflation levels, the traction force increased by almost twice as much (about 2 N) as the uninflated case. However, the over-inflation of the chambers can lead to an increment of the friction between tracks and chambers and result in the robot getting stuck, or to an excessive stretch of the membrane.

\section*{CONCLUSIONS AND DISCUSSION}
\justifying 
In this paper, we present the SoftSCREEN system, a novel robotic device for colonoscopy. The proposed design derives from the analysis of state-of-the-art of the locomotion strategies and shape reconfigurability adopted for the navigation inside the colon.
The experimental tests provided a preliminary insight to demonstrate the navigation capabilities of our system when moving through irregular, curved and deformable lumen.
Regardless the expansion of the chambers, in the case of a collapsed phantom, the robot was always able to open the occluded membrane thanks to the everting motion of the tracks that contributes to open the lumen automatically, by guiding the colon lining towards the outside of the periphery of the robot.
The use of the inflating chambers has been validated to reconfigure the shape of the robot, thus providing optimised visual inspection of the inside of the lumen investigated, and to increase the traction force exerted by the tracks. The authors believe that the controlled inflation of the chambers could be used in the context of colonoscopy to locally distend the folds of the linen by stretching the membrane surrounding the robot, improving local visualization of the thus reducing the missing rate of polyps without the need of typically painful insufflation. Moreover, the independent control of the chambers inflation level can enable camera tilting as well as steering of potential tools, e.g. snaps and needles for biopsies.

In conclusion, the experimental tests validated the principle of motion of the large-scaled proof-of-concept prototype and, although different from \textit{in vivo} or \textit{ex vivo} testing in colon, the system has shown great navigation capabilities in silicone-based colon phantoms, succeeding in navigating all tested scenarios and showing reliable locomotion. 

Future work will focus on the miniaturization of the robot to match the dimension of the human colon, and the miniature robot will be envisioned as a mobile module for the integration of a single front camera, similar to the one used in standard colonoscopes. The integration of surgical tools will be also considered. Moreover, electronic components such as motor and camera will be protected from the external environment with sealing and the design of the movable parts, such as the expandable mechanism here presented, will be further optimised to ensure safety in the interaction with the human colon. Finally, the evaluation of locomotion capabilities of the tracks with respect to the radial reconfiguration of the miniature robot, as well as the visual output provided by the single camera, will need to be conducted in more realistic scenarios, such \textit{ex vivo} and \textit{in vivo} porcine colons.



\section*{ACKNOWLEDGMENT}
This research was funded in whole, or in part, by the
Wellcome/EPSRC Centre for Interventional and Surgical
Sciences (WEISS) [203145/Z/16/Z]; the Engineering and
Physical Sciences Research Council (EPSRC)
[EP/P027938/1, EP/R004080/1, EP/P012841/1]; the Royal
Academy of Engineering Chair in Emerging Technologies
Scheme [CiET1819/2/36] and by the Rosetrees
Trust/Stoneygate Trust Enterprise Fellowship Scheme
[M884]. For the purpose of open access, the authors have
applied a CC BY public copyright licence to any author
accepted manuscript version arising from this submission.

\nocite{*}
\bibliographystyle{IEEEtran}
\bibliography{RAL}

\end{document}